\PassOptionsToClass{balance=false}{acmart}
\documentclass[sigplan,screen,nonacm]{acmart}
\usepackage{array,tabularx,longtable,multirow}
\usepackage{enumitem}
\usepackage{placeins}
\usepackage{algorithm}
\usepackage{algpseudocode}
\usepackage{xurl}

\newcommand{\sysname}{\textsc{ReWY}}
\theoremstyle{acmplain}
\newtheorem{theorem}{Theorem}[section]

\newtheorem{corollary}[theorem]{Corollary}

\theoremstyle{acmdefinition}

\DeclareMathOperator{\diag}{diag}

\newcommand{\low}{\mathcal L_{<}}
\newcommand{\causal}{\mathcal L_{\le}}
\newcommand{\RR}{\mathbb R}
\newcommand{\ip}[2]{\left\langle #1,#2\right\rangle_F}
\newcommand{\Had}{\odot}
\newcommand{\invHad}{\oslash}

\newcommand{\smallurl}[1]{\url{#1}}

\title[ReWY: Faster Backpropagation for Delta-Rule Attention]{\sysname{}: Faster Backpropagation for Delta-Rule Attention \texorpdfstring{---}{—} From 7 to 3 Matrix Multiplications}

\author{Yi Zhang}
\affiliation{%
  \institution{Tianjin University}
  \city{Tianjin}
  \country{China}}
\email{zhang\_e@tju.edu.cn}

\author{Zhixin Zhao}
\affiliation{%
  \institution{Tianjin University}
  \city{Tianjin}
  \country{China}}
\email{zhao612@tju.edu.cn}

\author{Weiwei Sun}
\affiliation{%
  \institution{Tianjin University}
  \city{Tianjin}
  \country{China}}
\email{sww@tju.edu.cn}

\author{Yitao Hu}
\authornote{Corresponding author.}
\affiliation{%
  \institution{Tianjin University}
  \city{Tianjin}
  \country{China}}
\email{yitao@tju.edu.cn}
\hypersetup{pdfcontactemail={yitao@tju.edu.cn}}

\author{Sheng Chen}
\affiliation{%
  \institution{Tianjin University}
  \city{Tianjin}
  \country{China}}
\email{chensheng@tju.edu.cn}

\author{Keqiu Li}
\affiliation{%
  \institution{Tianjin University}
  \city{Tianjin}
  \country{China}}
\email{keqiu@tju.edu.cn}

\begin{document}
\begin{abstract}
Delta-rule linear attention underlies language-model architectures such as Kimi Linear and Qwen3-Next. Chunkwise implementations expose matrix parallelism, but their backward passes compute auxiliary adjoints that the final input gradients do not need. We present \sysname{}, an algebraic reformulation that reduces this work before optimizing accelerator execution. By differentiating the chunk correction as one triangular residual system, \sysname{} eliminates the inverse adjoint and reduces seven local WY matrix products to three. Two complementary transformations expose further reuse: centered matrix contractions replace pairwise processing of vector gates, and scalar-gate reductions reuse key-gradient products instead of forming a key Gram matrix. Together, these identities simplify the backward graph for DeltaNet, GDN, KDA, and GDN2 while preserving the recurrence and its gradients in exact arithmetic, without changing the model architecture or training objective.

We implement the reduced graph on GPUs and Ascend NPUs with schedules that balance reuse and storage. Under matched recomputation and full-saving policies, precision-qualified six-shape comparisons with frozen FLA references yield 1.15--1.81$\times$ geometric-mean complete low-level backward speedups on NVIDIA GPUs. All four operators also achieve 1.18--1.70$\times$ on NVIDIA H100 and NVIDIA H200 while retaining the original FP32 working matrices. A separate public-autograd evaluation on Ascend 950DT, including gradient conversions, yields 1.09--2.89$\times$ under the available matched policies. Paired ablations establish additional \sysname{} gains after optimizing the complementary contractions. Complete-gradient checks validate the recorded operator cases; supplementary full-model backward measurements evaluate integration into community FLA models and an identified GDN2 layer port. These results show how simplifying an existing operator's derivative graph translates into accelerator performance.
Our implementation and reproduction materials are available at \url{https://github.com/flashserve/ReWY}.
\end{abstract}

\maketitle
\section{Introduction}
\label{main:intro}
Linear attention has become a practical building block of industrial language-model architectures. Kimi Linear uses Kimi Delta Attention (KDA) in an approximately 3:1 mixture with global Multi-Head Latent Attention~\cite{kda,kimi_code}. Qwen3-Next places Gated DeltaNet (GDN) in 36 of its 48 layers, and the Qwen3.5-0.8B text backbone uses it in 18 of 24 layers~\cite{qwen_next,qwen35}. These models repeatedly invoke the same family of delta-rule operators to update a recurrent matrix state. Optimizing that primitive therefore addresses an existing model-computation workload shared across many layers and architectures.

The attraction of these operators is their sequence-length-independent recurrent state and linear sequence complexity. Training, however, must also expose enough parallel work to exploit accelerator matrix units. Chunkwise algorithms express causal interactions within each chunk as dense matrix operations and connect chunks through a state scan~\cite{deltanet,gdn,kda,gdn2}. Flash Linear Attention (FLA) provides widely used implementations of this computation; the Kimi authors release their KDA kernel through FLA itself~\cite{fla,kimi_code}. The systems question is how much work this core operator actually needs to perform, particularly in its backward pass.

\begin{figure*}[t]
\centering
\includegraphics[width=\linewidth]{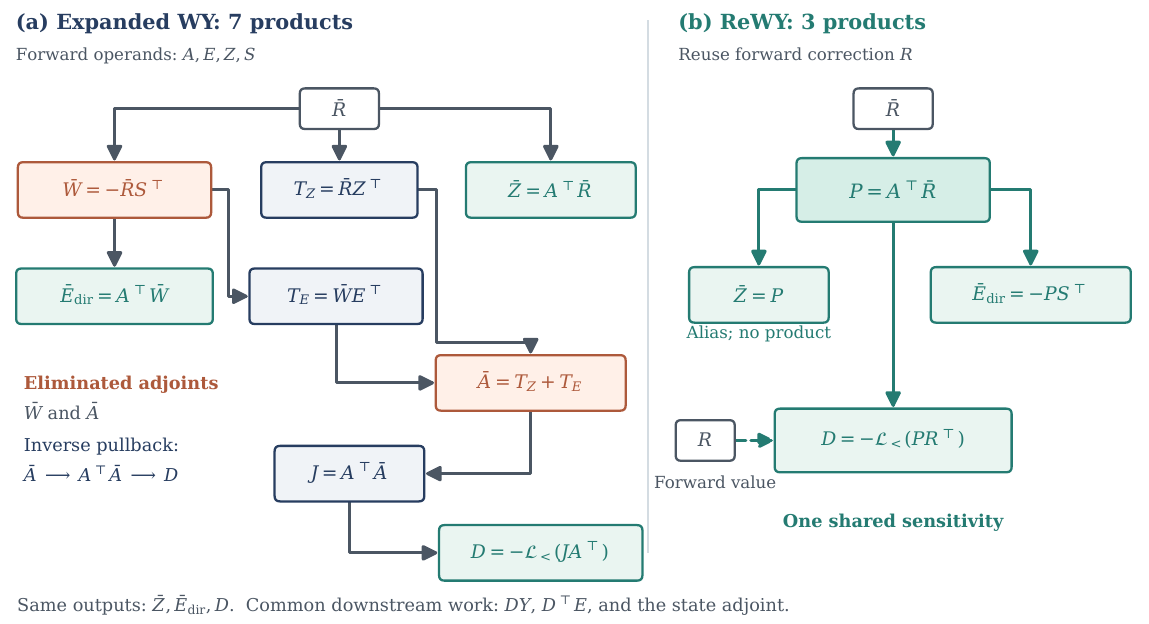}
\Description{Two dependency graphs compare the expanded WY backward with the residual backward. The expanded graph contains seven matrix products; the residual graph contains three and reuses the forward correction R.}
\caption{The local backward computation graph before and after \sysname{}. Arrows show dependencies between computed values; forward operands appear in the node formulas. The expanded graph forms \(\bar W\) and \(\bar A\) (orange) and propagates the inverse adjoint through two matrix products. \sysname{} shares \(P\) across the same three required gradients (green), using the forward correction \(R\). The graphs contain seven and three matrix products, respectively; addition, masking, and the alias \(\bar Z=P\) are not counted as products.}
\label{fig:overview}
\end{figure*}

\paragraph{A redundant derivative graph.}
The conventional WY decomposition is useful for parallelizing the forward pass, but differentiating its intermediate operations separately creates avoidable backward work. It forms adjoints of two auxiliary objects---a transformed key matrix and a triangular inverse---only to multiply them again into the gradients that the caller needs. Fusion and tiling can accelerate these products, but executing the same derivative graph still pays for their arithmetic and intermediate lifetimes. At chunk size 64 and key/value width 128, the local WY graph contains seven matrix products and 7.34 million dense FLOPs per chunk/head, before the common state scan and interaction gradients are counted.

\paragraph{Eliminate work, then schedule what remains.}
Our key observation is that the chunk correction satisfies one triangular residual equation. Differentiating this composite map exposes a shared reverse sensitivity and makes the inverse-adjoint chain unnecessary. We call the resulting algorithm \sysname{}. As shown in Figure~\ref{fig:overview}, \sysname{} reduces the local WY computation from seven products to three, removing 42.9\% of its dense FLOPs at the above dimensions. The forward recurrence, attention semantics, and exact gradients remain unchanged. This is an operator algorithm change applicable to DeltaNet, GDN, KDA, and GDN2; it does not introduce a new neural-network architecture.

The reduced graph also exposes reuse in the remaining interactions. \emph{Matrix intra} factors vector-gated pairwise interactions into centered matrix contractions, replacing repeated token-pair processing with matrix-unit work. \emph{Gram elimination} computes scalar-gate reductions from key-gradient products that are already needed, avoiding a separate key Gram matrix. These transformations complement \sysname{}: the latter simplifies the coefficient producer, while the former optimize its gradient consumers. Their independent and conditional gains are measured separately.

Turning this algebra into speed requires an execution strategy. Sharing the reverse sensitivity can save products while adding memory traffic; fusing all consumers can instead extend live ranges and restrict occupancy. Our implementations select joint or separate producer--consumer schedules by device and operator, preserve existing FP32 working matrices and accumulations, and use one schedule across the evaluated shapes. Matched saving policies prevent existing FLA activation-saving support from being counted as a new algorithmic contribution. The evaluation measures the complete backward operator, rather than extrapolating a local kernel improvement to the whole computation.

\paragraph{Contributions.}
This paper makes three contributions:
\begin{itemize}
\item \textbf{A smaller delta-rule derivative graph.} \sysname{} eliminates four local matrix products and the inverse adjoint while recovering all input, state, and gate gradients. We give the resulting formulas in the main text and the full equivalence proof in Appendix~\ref{sec:residual}.
\item \textbf{Contractions and schedules.} Centered matrix intra and Gram elimination remove distinct sources of repeated work. GPU and NPU schedules realize this reuse while controlling shared storage, numerical range, and intermediate lifetimes.
\item \textbf{Complete-operator evaluation.} Across six shapes, precision-qualified NVIDIA GPU comparisons yield complete backward speedups of 1.15--1.81\(\times\). A separate Ascend 950DT public autograd study yields 1.09--2.89\(\times\), with standalone residual and FP32 matrix-intra controls. Ascend 910C comparisons retain their stated precision qualifications. Paired Gram controls establish additional \sysname{} gains. Tests check every gradient, and separate allocation measurements characterize storage.
\end{itemize}

Our implementation and reproduction materials are available at \url{https://github.com/flashserve/ReWY}.

The layer fractions above motivate the operator's relevance; they are not runtime fractions. Supplementary A800, Ascend 910C, and scoped H100/H200 full-model backward measurements in Section~\ref{sec:e2e} quantify the latter using FLA's existing full models and an identified GDN2 layer port. For example, the core backward accounts for 25.7--31.3\% of the unmodified FLA KDA model's measured backward interval across its input-size sweep. The paper's main performance claim remains the complete-operator improvement; model integration measures how that improvement carries through the surrounding computation.
\FloatBarrier

\section{Background}
\label{main:problem}
\subsection{Accelerator execution and chunkwise training}
GPU and NPU kernels combine matrix operations with elementwise transformations and reductions. A sequence of kernels exchanges intermediate arrays through device memory, whereas a fused kernel can reuse values in registers or on-chip storage. Fusion also keeps more operands and accumulators live. For chunkwise delta-rule operators, the execution schedule must balance this local reuse against the parallelism available across chunks, heads, and feature tiles.

Chunkwise computation separates two kinds of dependency. Within a chunk, triangular coefficient matrices encode the causal token interactions and expose matrix parallelism. Between chunks, a state scan propagates the recurrent memory. The backward pass has the same structure in reverse: a scan accumulates gradients from later chunks, after which local kernels compute the input gradients. Recomputing forward intermediates or saving them changes the work surrounding these local kernels. We compare the two policies separately.

\subsection{Delta-rule operators}
Consider one chunk of \(n\) tokens and one attention head. The incoming memory is \(S\in\RR^{d_k\times d_v}\), queries and keys are \(Q,K\in\RR^{n\times d_k}\), and values are \(V\in\RR^{n\times d_v}\). Token vectors are columns, and the corresponding matrices stack their transposes as rows. A fixed query scale is absorbed into \(Q\). An overbar denotes a reverse-mode adjoint, with outer normalization and gate activations handled by the chain rule.

Let \(g_i\) be the log-decay vector, \(b_i\) the erase gate, and \(w_i\) the write gate. The four operators use the recurrence
\begin{align*}
 S_i^-&=\diag(\exp g_i)S_{i-1},\\
 r_i&=w_i\Had v_i-(S_i^-)^\top(b_i\Had k_i),\\
 S_i&=S_i^-+k_i r_i^\top,\qquad o_i=S_i^\top q_i.
\end{align*}

The decayed state \(S_i^-\) predicts a value from the gated key. The correction \(r_i\) subtracts this prediction from the gated input value, and the rank-one update writes the correction along \(k_i\). The output reads the updated state, including the current write. Table~\ref{tab:family} specifies the gate parameterizations; \(\Had\) denotes elementwise multiplication.

\begin{table}[htbp]
\footnotesize
\centering
\caption{Gate structures in the common recurrence. Scalar gates are broadcast across features.}
\label{tab:family}
\small
\setlength{\tabcolsep}{3pt}
\begin{tabular}{@{}llll@{}}
\toprule
Operator & Log decay \(g_i\) & Erase \(b_i\) & Write \(w_i\)\\
\midrule
DeltaNet & \(0\) & \(\beta_i\mathbf1_{d_k}\) & \(\beta_i\mathbf1_{d_v}\)\\
GDN & \(\eta_i\mathbf1_{d_k}\) & \(\beta_i\mathbf1_{d_k}\) & \(\beta_i\mathbf1_{d_v}\)\\
KDA & vector & \(\beta_i\mathbf1_{d_k}\) & \(\beta_i\mathbf1_{d_v}\)\\
GDN2 & vector & independent vector & independent vector\\
\bottomrule
\end{tabular}
\end{table}

\subsection{The chunk system and conventional WY backward}
Let \(R\in\RR^{n\times d_v}\) stack the corrections, \(G_i=\sum_{t\le i}g_t\) be cumulative log decay, and \(\gamma=\exp G\). Define
\begin{align*}
 E&=B_e\Had K\Had\gamma,& Y&=K\invHad\gamma,\\
 Z&=W_e\Had V,& Q_g&=Q\Had\gamma.
\end{align*}

where \(B_e,W_e\) stack the erase/write gates and \(\invHad\) denotes elementwise division. The factors \(E\) and \(Y\) account for decay before an erase and after an earlier write. Substituting the unrolled state gives
\begin{equation*}
 R_i=Z_i-E_iS-\sum_{j<i}(E_iY_j^\top)R_j.
\end{equation*}
In matrix form,
\begin{equation*}
 \begin{aligned}
 MR&=Z-ES,\\
 M&=I+L,\qquad L=\low(EY^\top).
 \end{aligned}
\end{equation*}

The projection \(\low\) retains the strictly lower triangle. Thus \(M\) is unit lower triangular, and each coefficient describes how an earlier correction affects a later prediction. Appendix~\ref{app:forward} derives this system for all compatible dimensions.

Writing \(A=M^{-1}\), the conventional WY implementation forms \(W=AE\) and \(U=AZ\), then computes \(R=U-WS\) during the state scan. This organization allows the local factors \(A,W,U\) to be prepared independently across chunks. The token outputs and outgoing state are
\begin{equation*}
 O=Q_gS+C_qR,\qquad S^+=\Lambda S+F^\top R,
\end{equation*}

where \(\Lambda=\diag(\gamma_n)\), \(F=Y\Lambda\), and \(C_q=\causal(Q_gY^\top)\). The inclusive projection \(\causal\) accounts for the current write in each output.

Given an accumulated correction adjoint \(\bar R\), the expanded WY backward forms \(\bar W=-\bar RS^\top\), then \(\bar A=\bar RZ^\top+\bar WE^\top\), and finally propagates \(\bar A\) through the inverse as \(-A^\top\bar A A^\top\). Together with \(\bar Z=A^\top\bar R\) and \(\bar E_{\rm dir}=A^\top\bar W\), this local graph contains seven matrix products. It materializes \(\bar W\) and \(\bar A\) even though the downstream computation ultimately requires the coefficient and input gradients. This is the graph simplified by \sysname{}.

\section{\sysname{}: Algorithm and Analysis}
\label{main:method}
The algorithm simplifies the local backward graph in two steps. \sysname{} first removes the inverse-adjoint chain and exposes a shared sensitivity \(P\). Matrix intra and Gram elimination then reuse the resulting coefficient adjoint to compute vector- and scalar-gate gradients, respectively. The state scan supplies the inter-chunk adjoints, while device schedules determine where the local intermediates are consumed.

\subsection{Differentiating the residual equation}
\label{main:residual}
The chunk correction obeys \(MR=Z-ES\), where \(M=I+L\) and \(A=M^{-1}\). Rather than materializing derivatives of the forward auxiliaries, \sysname{} computes a shared sensitivity \(P=A^\top\bar R\). This uses the classical reverse derivative of a linear solve~\cite{giles}; the contribution is its application to the composite delta-rule graph and the resulting elimination of intermediate adjoints, not a new general differentiation rule. The following result gives the complete local gradient interface. Appendix~\ref{sec:residual} proves equivalence to the expanded graph.

\begin{samepage}
\begin{theorem}[\sysname{} pullback]
\label{thm:residual}
For strictly lower triangular \(L\), let \(M=I+L\) and \(R\) satisfy \(MR=Z-ES\). Given any adjoint \(\bar R\), the sensitivity \(P=M^{-\top}\bar R\) yields
\begin{align*}
 \bar Z&=P,\qquad \bar E_{\mathrm{dir}}=-PS^\top,\\
 \bar S_{\mathrm{res}}&=-E^\top P,\\
 D:=\bar L&=-\low(PR^\top).
\end{align*}
When \(L=\low(EY^\top)\), its remaining contributions are
\begin{equation*}
 \bar E_{\mathrm{intra}}=DY,\qquad
 \bar Y_{\mathrm{res}}=D^\top E.
\end{equation*}
\end{theorem}
\end{samepage}

Thus the complete erase-factor gradient is \(-PS^\top+DY\). The three new products are \(A^\top\bar R\), \(PR^\top\), and \(PS^\top\); the interaction contractions and state scan are shared with the reference algorithm.

\begin{figure}[htbp]
\centering
\includegraphics[width=\linewidth]{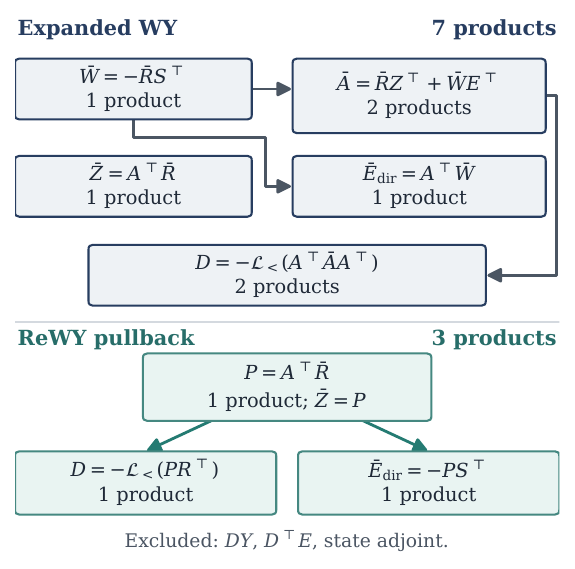}
\Description{The expanded WY pullback uses seven matrix products and auxiliary adjoints for W and A. The residual pullback uses three products, sharing P between the Z, E, and coefficient gradients.}
\caption{Local WY pullback before and after \sysname{}. Box labels count dense matrix products: seven in the expanded graph and three in the reduced graph. The shared intra contractions and state adjoint are excluded from both counts.}
\label{fig:dataflow}
\end{figure}

\paragraph{State and gate composition.}
The local theorem is applied after accumulating both uses of the correction:
\begin{equation*}
 \bar R=C_q^\top\bar O+F\bar S^+.
\end{equation*}

Both output and outgoing-state contributions are accumulated before the local pullback. \sysname{} leaves the reverse-state scan unchanged; the scan supplies the incoming-state gradient once. Shared scalar erase/write gates combine both paths using an undivided expression at zero or tiny beta. GDN2 retains independent erase/write gradients. Appendix~\ref{sec:complete} and Corollary~\ref{cor:beta} derive these identities; Appendix~\ref{app:algorithm} gives the full reverse algorithm.

\begin{algorithm}[htbp]
\small
\caption{\sysname{} local pullback}
\label{alg:residual}
\begin{algorithmic}[1]
\Require \(A=M^{-1}\), factors \(E,Y\), incoming state \(S\), correction \(R\), accumulated adjoint \(\bar R\).
\Ensure contributions to \(\bar Z,\bar E,\bar Y,\bar S\).
\State Compute the shared sensitivity \(P\gets A^\top\bar R\).
\State Compute \(D\gets-\low(PR^\top)\), retaining the strict causal mask.
\State Set \(\bar Z\gets P\) and \(\bar E_{\rm dir}\gets-PS^\top\).
\State Add the interaction terms: \(\bar E\gets\bar E_{\rm dir}+DY\), \(\bar Y\gets D^\top E\).
\State Supply \(\bar S_{\rm res}\gets-E^\top P\), or use the equivalent scan term \(-W^\top\bar R\), once.
\end{algorithmic}
\end{algorithm}

\subsection{Arithmetic and intermediate storage}
The expanded graph has one product for \(\bar W\), two for \(\bar A\), two for the inverse pullback, and one each for \(\bar Z\) and \(\bar E_{\rm dir}\). \sysname{} computes \(P\), \(PR^\top\), and \(PS^\top\). Figure~\ref{fig:dataflow} compares the dependencies, and Algorithm~\ref{alg:residual} shows how the reduced graph supplies the common downstream gradients. The seven-product count describes the expanded frozen-reference graph; it is not a claim that every optimized implementation performs seven products. The inspected FlashQLA revision already combines the residual right-hand side but retains the two-sided inverse pullback~\cite{flashqla}. This paper does not measure performance against that implementation.

The dense arithmetic removed per chunk/head is
\begin{equation*}
 \boxed{\Delta\mathcal F=4n^2d_k+4n^3.}
\end{equation*}
At \(n=64\), \(d_k=d_v=128\), the local WY cost falls from 7.34 to 4.19 million FLOPs (42.9\%). This count excludes the unchanged state scan, common interaction products, and activation recomputation; it is not a prediction of complete-backward speedup. Appendix~\ref{app:residual_cost} details the cost.

The eliminated adjoints have shapes \(\bar A\in\RR^{n\times n}\) and \(\bar W\in\RR^{n\times d_k}\). The new sensitivity \(P\in\RR^{n\times d_v}\) feeds value, direct-state, and coefficient gradients, creating a choice between local reuse and shared storage. Fusing its production and consumption can keep \(P\) on chip; separate kernels exchange it or the resulting coefficient matrices through device memory. The resulting allocation peak is set by the overlap of these lifetimes. Section~\ref{sec:memory_attribution} measures that peak for the same complete implementations used in the timing comparison.

\subsection{Matrix intra for vector-gated interactions}
\label{main:matrixintra}
\sysname{} produces the causal coefficient adjoint \(D\); the output path produces \(H_q=\causal(\bar OR^\top)\). Both must be propagated to query, key, and gate gradients. With vector decay, each token pair carries a feature-dependent factor \(\exp(G_{ik}-G_{jk})\). Evaluating this factor inside each pairwise interaction obscures the matrix structure of the contraction.

GLA and KDA establish separable cumulative-gate factors and stability-aware secondary chunking~\cite{gla,kda}. We use this structure to reorganize the four backward contractions and their feature-local gradients; separability alone is not a new result. The gate ratio factors into \(\exp(G_{ik})\) and \(\exp(-G_{jk})\). Scaling the sending features before multiplication and the receiving features afterward yields
\begin{equation}
 \begin{aligned}
 U_q&=H_qY,\qquad U_e=DY,\\
 U_y&=H_q^\top Q_g+D^\top E.
 \end{aligned}
 \label{main:intra}
\end{equation}

and the physical input gradients
\begin{align}
 \bar Q_{\mathrm{intra}}&=\gamma\Had U_q, \notag\\
 \bar K_{\mathrm{intra}}&=\gamma\Had B_e\Had U_e+\gamma^{-1}\Had U_y, \notag\\
 \bar B_{e,\mathrm{intra}}&=\gamma\Had K\Had U_e, \notag\\
 \bar G_{\mathrm{intra}}&=Q_g\Had U_q+E\Had U_e-Y\Had U_y.
 \label{main:intraoutputs}
\end{align}

Here \(\gamma^{-1}\) denotes elementwise reciprocals. For example, the \((i,k)\) entry of \(\gamma\Had H_qY\) equals \(\sum_j(H_q)_{ij}K_{jk}\exp(G_{ik}-G_{jk})\). The other paths follow the same factorization or its transpose. The matrix-intra stage therefore consists of four matrix products and feature-local chain rules, with the causal masks already included in \(H_q,D\).

\paragraph{Centered factors.}
Separately forming the exponential factors can create large intermediate values even when their ratios are moderate. For a feature-wise center \(c\), let \(C_c=\diag(\exp c)\) and use
\begin{equation}
 (\widetilde Q_g,\widetilde E,\widetilde Y)
 =(Q_gC_c^{-1},\,EC_c^{-1},\,YC_c).
 \label{main:diagonalaction}
\end{equation}

The factors \(C_c\) and \(C_c^{-1}\) cancel in the coefficient products and in each completed input gradient. Thus the centered factors replace \(\gamma\) and \(\gamma^{-1}\) by \(\alpha=\exp(G-c)\) and \(\zeta=\exp(c-G)\), respectively. Appendix~\ref{sec:intra} proves the factorization and scaling invariance.

Let \(m_k\) and \(M_k\) be the minimum and maximum of \(G_{ik}\) over tokens, and set \(\Delta_k=M_k-m_k\). The midpoint minimizes the largest centered log magnitude:
\begin{equation*}
 \begin{gathered}
 c_k^*=\frac{M_k+m_k}{2},\\
 \min_{c_k\in\RR}\max_i|G_{ik}-c_k|=\frac{\Delta_k}{2}.
 \end{gathered}
\end{equation*}

Consequently, the exponential factors satisfy
\begin{equation*}
 \begin{aligned}
 &\max_i\{\exp(G_{ik}-c_k^*),\exp(c_k^*-G_{ik})\}\\
 &\qquad\leq \exp(\Delta_k/2).
 \end{aligned}
\end{equation*}

The implementation centers feature tiles and checks their gate ranges. Stronger gates use a validated pairwise or subchunk path. These safeguards accompany the algebraic factorization.

Matrix intra retains the \(O(n^2d_k)\) interaction count but groups it into regular matrix operations with fused feature-local derivatives. The frozen KDA implementation already uses matrix multiplication for part of its intra computation; our consumer reorganizes whole-chunk contractions, centering, and gradient production. Since \(D\) can come from either WY backward graph, the matrix-intra contribution can be evaluated separately from residual elimination.

\subsection{Gram elimination for scalar-gate gradients}
\label{main:gram}
For scalar-gated GDN, the remaining gate reduction has a different redundancy. The key gradient already contracts the coefficient adjoint with the keys, but an expanded gate path also forms a key Gram matrix. The gate gradient needs only weighted row and column sums of that matrix, which can be obtained from the existing contractions.

The scalar-gated coefficient is
\begin{equation*}
 L_{ij}=\mathbf1_{\{i>j\}}\beta_i
          \exp(G_i-G_j)\langle K_i,K_j\rangle.
\end{equation*}

Let \(D=\bar L\) include the strict mask, set \(\Gamma_{ij}=D_{ij}\exp(G_i-G_j)\), and define
\begin{equation*}
 J=\diag(\beta)\Gamma\Had(KK^\top).
\end{equation*}

Differentiating the receiving and sending positions of the log gate yields
\begin{equation*}
 \bar G^{(L)}=J\mathbf1_n-J^\top\mathbf1_n.
\end{equation*}

Meanwhile, the key-gradient path computes
\begin{align*}
 A_K&=\Gamma K,\qquad B_K=\Gamma^\top(\diag(\beta)K),\\
 \bar K^{(L)}&=\diag(\beta)A_K+B_K.
\end{align*}

Interchanging the finite token and feature sums gives
\begin{equation*}
 \begin{aligned}
 \sum_jJ_{ij}&=\beta_i\langle K_i,(A_K)_i\rangle,\\
 \sum_jJ_{ji}&=\langle K_i,(B_K)_i\rangle.
 \end{aligned}
\end{equation*}
The gate derivative is therefore
\begin{equation*}
 \begin{aligned}
 \bar G_i^{(L)}&=\beta_i\langle K_i,(A_K)_i\rangle\\
 &\quad-\langle K_i,(B_K)_i\rangle.
 \end{aligned}
\end{equation*}

The beta contribution uses the same contraction:
\begin{equation*}
 \bar\beta_i^{(L)}=\langle K_i,(A_K)_i\rangle.
\end{equation*}

This expression is valid at zero beta. The local gate contribution is added to the other cumulative-gate derivatives before the reverse sum. Appendix~\ref{sec:gram} gives the rectangular bilinear identity and its proof.

Given \(A_K,B_K\), the replacement costs \(O(nd_k)\) elementwise operations and reductions, eliminating a dense \(KK^\top\) product of approximately \(2n^2d_k\) FLOPs and its \(n\times n\) intermediate. The transformation applies when the pairwise matrix is consumed through these reductions. Scalar GDN has this structure; vector-gated KDA/GDN2 instead use the feature-dependent matrix-intra factorization. Gram elimination accepts the coefficient adjoint from either WY formulation, providing a second independent component control.

\subsection{Execution and numerical precision}
\label{main:execution}
The transformed graph has a common producer--consumer interface across the four operators. After the reverse-state scan, a producer computes \(P\), value/write gradients, and \(D\). Consumers propagate the output and residual coefficients to query, key, and gate gradients. The final reverse cumulative sum converts cumulative-gate gradients to token-local gradients. This separation exposes both matrix reuse within a chunk and independent work across chunks and heads.

\paragraph{Joint and separate consumers.}
A joint kernel consumes the sensitivity and state-direct terms together with the intra contractions, reducing intermediate transfers. A separate schedule shortens live ranges and gives each stage its own task partition. NVIDIA RTX 5070Ti uses a joint KDA consumer with streamed value tiles. The A800 DeltaNet/GDN paths consume \(P\) locally without a new global buffer. On Ascend 910C, GDN separates the scalar state-endpoint reduction and gate scan into a Vector kernel. The FP32 vector-gate schedules on Hopper and 950 are specified with their measurements in Sections~\ref{sec:h100} and~\ref{sec:ascend950}. Appendix~\ref{app:execution} details the schedules; Appendix~\ref{app:precision_status} identifies precision exclusions.

The state scan remains serial across chunks and parallel across independent value columns. Candidate schedules are fixed per device across the measured shapes. The A800 DeltaNet state kernel retains the original K64 arithmetic with three pipeline stages; GDN retains the original state autotuner. The frozen references keep their native device autotuning, including the multi-stage A800 search space.

\paragraph{Precision and numerical stability.}
\label{main:numeric}
The main experiments use BF16 features, effective erase/write weights, and output adjoints, with FP32 log gates and state interfaces. Our precision contract preserves existing operand precision, stored matrix formats, and FP32 accumulation; exclusions are specified in Appendix~\ref{app:precision_status}. Newly shared \(P\) storage follows the consumer requirements: RTX 5070Ti uses the input format for DeltaNet/GDN/KDA, while GDN/NPU retains FP32 values for its value/beta consumers. Table~\ref{tab:gpu_precision} specifies these uses.

Centered matrix intra uses gate-range dispatch, and small shared beta uses the undivided derivative in \eqref{eq:betadirect}. Independent GDN2 erase/write gates retain separate derivatives. These choices are checked against the complete-gradient thresholds in Section~\ref{main:eval}; partitions, masks, and shared-parameter reductions are derived in Appendix~\ref{app:generality}.

\section{Experiments}
\label{main:eval}
We evaluate complete-backward latency and memory across four delta-rule operators on NVIDIA GPUs and Ascend NPUs. Section~\ref{sec:ascend950} specifies the public-interface timing boundary for Ascend 950DT; Section~\ref{sec:h100} reports H100/H200 results. Component experiments separate the effect of \sysname{} from matrix intra and Gram elimination, connecting the local graph analysis to complete-operator performance.

\begin{figure*}[t]
\centering
\includegraphics[width=0.9\textwidth]{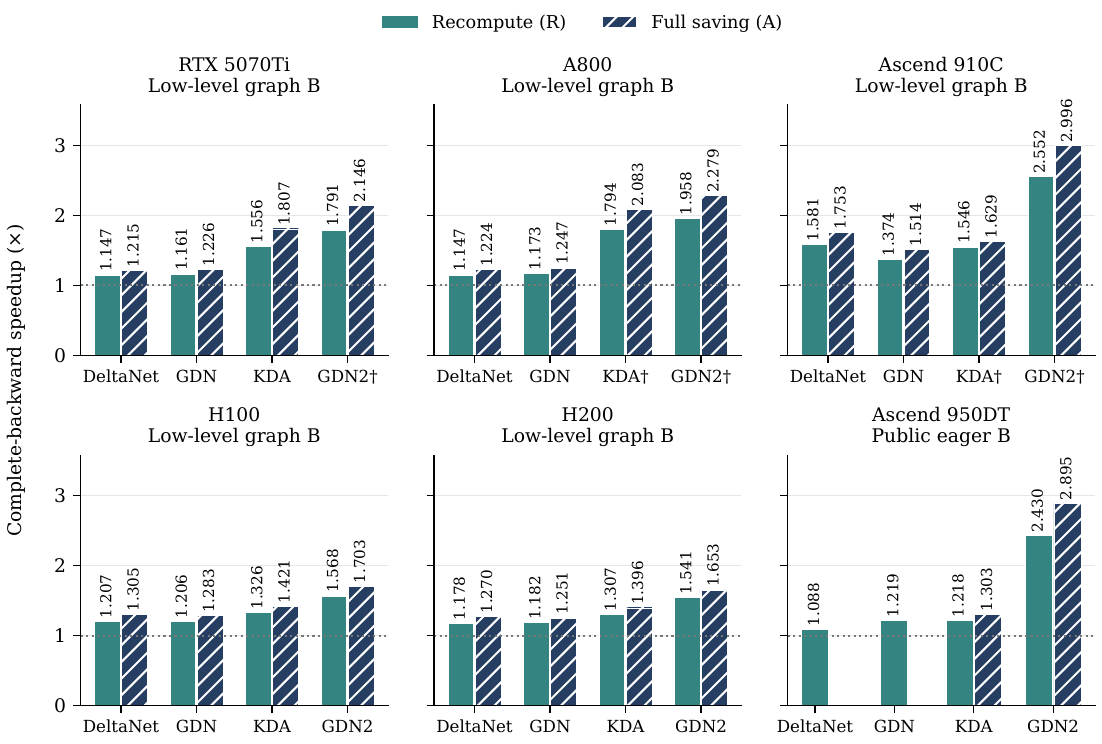}
\Description{Six panels show measured complete-backward speedups for DeltaNet, GDN, KDA, and GDN2. The top row, from left to right, shows H100, H200, and A800; the bottom row shows Ascend 950DT, Ascend 910C, and RTX 5070Ti. Policies R and A are shown where measured. Ascend 950DT uses public eager backward; the other panels use low-level graph backward. Daggers mark historical precision limitations.}
\caption{Complete-backward speedups: six-shape geometric means under matched recomputation (R) and full saving (A). The dotted line denotes parity. Only measured positive aggregate gains are drawn. The Ascend 950DT panel uses public eager backward including gradient conversions; other panels use low-level graph backward. Cohorts are not pooled. Daggers retain the historical precision limitation in Section~\ref{app:precision_status}. Exact values appear in Tables~\ref{tab:main}, \ref{tab:950combined}, and~\ref{tab:h100_operator}.}
\label{fig:backward_speedups}
\end{figure*}

\subsection{Experimental setup}
The primary latency and memory experiments use an RTX 5070Ti GPU (12\,GiB, SM120), an NVIDIA A800-SXM4-80GB GPU (80\,GB, SM80), and an Ascend 910C NPU (64\,GiB). The GPU environment is PyTorch 2.9.1+cu128 and Triton 3.5.1. The NPU environment is PyTorch 2.7.1, torch\_npu 2.7.1.post4, Triton 3.2.0, and CANN 9.0 beta1. Baselines are frozen FLA implementations with equivalent platform adaptations for the NPU; the reference revision and adaptations are listed in Appendix~\ref{app:evidence}.

\paragraph{Workloads and policies.}
We vary sequence length, batch size, and head count using six \((B,T,H)\) shapes:
\begin{equation}
 \begin{gathered}
 (1,512,4),\quad(1,2048,4),\quad(1,8192,4),\\
 (1,2048,16),\quad(2,2048,16),\quad(1,8192,16).
 \end{gathered}
 \label{main:shapes}
\end{equation}

The chunk size is \(C=64\), and \(d_k=d_v=128\). These shapes vary both the number of serial state updates and the number of independent head/chunk tasks. Each comparison uses either default recomputation (\(\mathrm R\)) or full saving (\(\mathrm A\)) on both sides. KDA uses FLA's existing saving switch; matched wrappers cache identical intermediates for the other interfaces. One validated implementation is fixed per operator/device across all six shapes and both policies.

\paragraph{Timing and allocation.}
Complete low-level backward (\(\mathrm B\)) includes all internal stages and required recomputation. Both sides exclude Python autograd overhead and the public wrapper's final gradient-type conversion; some optimized paths return FP32 where the reference returns BF16. Appendix~\ref{app:evidence} reports conversion-inclusive controls. Device-graph events measure warmed latency, with sixty observations per configuration on the GPUs and twenty on the NPU. Speedups are geometric means of the six reference/optimized median-latency ratios. Allocation measurements use warmed incremental peaks, include returned gradients, and report resident workspace separately.

\paragraph{Precision qualifications.}
Our precision contract preserves the reference input/output types, FP32 accumulations, and existing FP32 working matrices. Entries marked $\dagger$ cast existing matrix operands to lower precision and are reported only as diagnostic comparisons; they do not establish a gain under this contract. Appendix~\ref{app:precision_status} identifies these exclusions. Newly introduced factor storage is distinct from casting an existing matrix.

\subsection{Complete-backward latency}

Figure~\ref{fig:backward_speedups} shows positive aggregate gains for the retained comparisons under both saving policies. On RTX 5070Ti, R/A speedups are 1.15/1.22\(\times\) for DeltaNet, 1.16/1.23\(\times\) for GDN, and 1.56/1.81\(\times\) for KDA. On A800, DeltaNet and GDN achieve 1.15/1.22\(\times\) and 1.17/1.25\(\times\), respectively. These are complete-backward measurements, including the reverse-state scan and recomputation, with the reference autotuning retained.

Table~\ref{tab:main} reports one R/A pair per operator and device, with the precision exclusions above. Sections~\ref{sec:ascend950} and~\ref{sec:h100} report the FP32 Ascend 950DT and Hopper comparisons and their timing protocols.

Full saving produces larger aggregate speedups wherever both policies were measured. Removing common recomputation increases the fraction of backward time affected by the transformations, consistent with the complete-operator cost decomposition in Appendix~\ref{app:generality}. Within each pair, the ratio also varies with the workload. For example, A800 DeltaNet has minimum R/A ratios of 1.05/1.06\(\times\), and GDN has minima of 1.08/1.12\(\times\). Appendix~\ref{app:evidence} reports per-shape variation and the timing protocol.

\subsection{Backward memory footprint}
\label{sec:memory_attribution}
\begin{figure}[tbp]
\centering
\includegraphics[width=\linewidth]{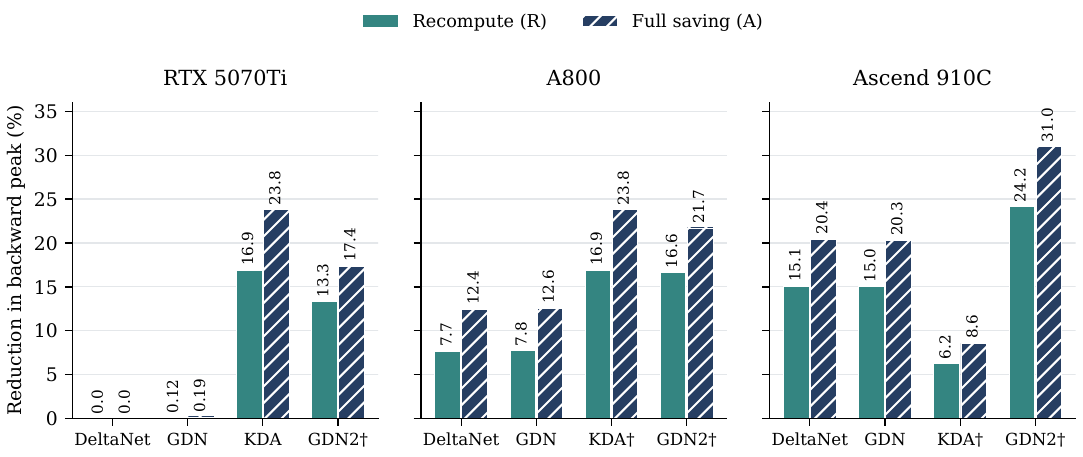}
\Description{Grouped bars show percentage reductions in incremental backward peak memory for four operators on three devices under recomputation and full saving. Daggers retain the historical precision qualifications. Exact values are in the numerical-value appendix.}
\caption{Reduction in incremental backward peak allocation, \(100(1-M_{\mathrm{opt}}/M_{\mathrm{ref}})\), at \(B=1,T=8192,H=16,d_k=d_v=128\). Implementations and dagger markers match Figure~\ref{fig:backward_speedups}. KDA/NPU additionally requires 16\,MiB of resident workspace per stream, excluded from these peaks.}
\label{fig:backward_memory}
\end{figure}

We measure the same complete implementations at \(B=1,T=8192,H=16,d_k=d_v=128\), with identical forward saved storage within each policy. Figure~\ref{fig:backward_memory} shows the reduction relative to each reference peak; Table~\ref{tab:space} gives absolute MiB values.

Allocation effects depend on the device schedule. On RTX 5070Ti, the DeltaNet peak is unchanged, whereas GDN saves 0.5\,MiB. KDA saves 130.50\,MiB under both policies. Its remaining R and A peaks are 641.50 and 417.50\,MiB, respectively. For A800, the savings are 32\,MiB for DeltaNet and 32.5\,MiB for GDN, with no new global \(P\) storage. Appendix~\ref{app:figure_values} lists the remaining peaks and their version qualifications. These differences reflect how shared sensitivity and downstream gradients overlap in memory.

Appendix~\ref{app:space} specifies saved storage, resident workspace, and allocation controls at two shapes. Peaks depend on the tested schedule.

\subsection{Transformation ablations}
The recorded RTX 5070Ti and Ascend 910C controls test each transformation alone and \sysname{} with a fixed companion. The companion is matrix intra for KDA and GDN2, and Gram elimination for GDN. Standalone results use their respective reference runs. Conditional gains use paired runs with the companion and its schedule fixed. The A800 results above correspond to the complete implementations. All recorded latency ablations are plotted; dagger markers identify historical GPU and Ascend 910C matrix-consumer controls whose precision status is detailed in Appendix~\ref{app:figure_values}.

\begin{figure}[htbp]
\centering
\includegraphics[width=\linewidth]{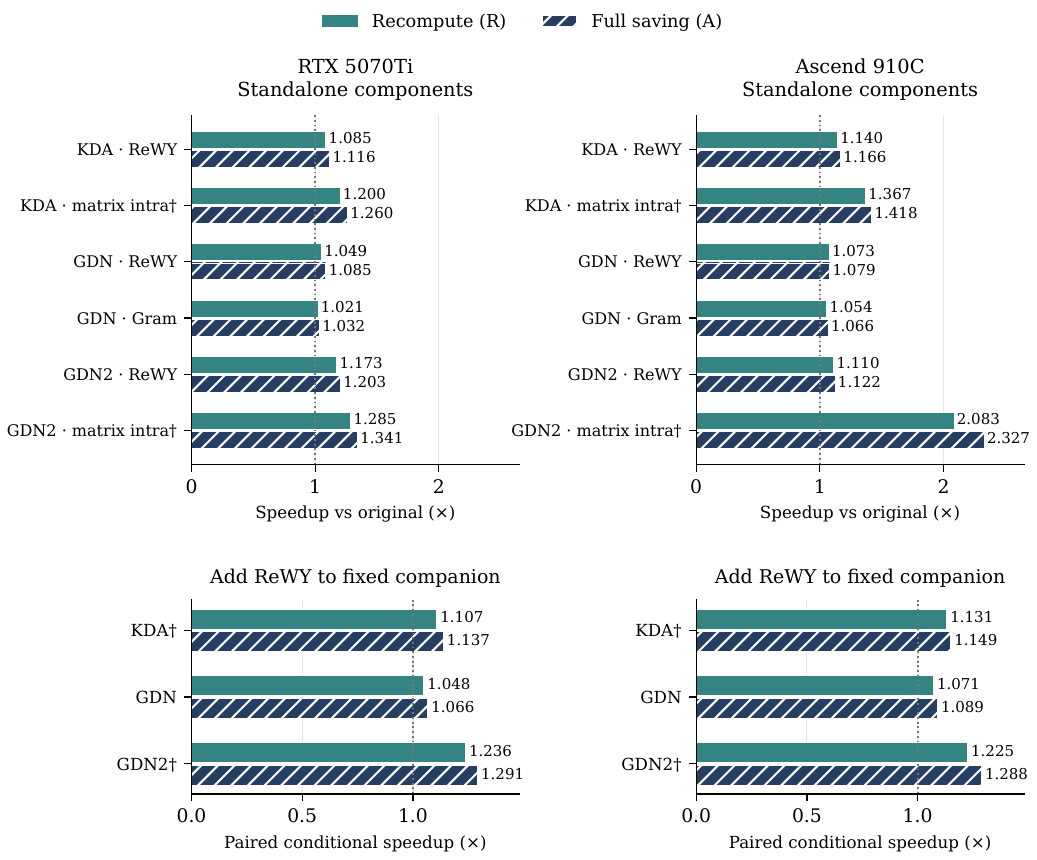}
\Description{Horizontal bar panels separate standalone residual and companion speedups from conditional gains obtained by adding the residual transformation to a fixed companion. Results cover KDA, GDN, and GDN2 on RTX and Ascend, with historical precision markers.}
\caption{Component speedups. Top: standalone \sysname{} and companion implementations. Bottom: paired gains from adding \sysname{} to a fixed companion. The companion is matrix intra for KDA/GDN2 and Gram elimination for GDN. Dagger-marked GPU and Ascend 910C matrix controls retain their recorded precision limitation. Standalone results use separate runs; conditional results use directly paired measurements (Tables~\ref{tab:ablation} and~\ref{tab:conditional}).}
\label{fig:component_speedups}
\end{figure}

\paragraph{Standalone and paired gains.}
The FP32 950 controls in Section~\ref{sec:ascend950} establish the independent effects of matrix intra and \sysname{} for KDA and GDN2. For GDN, Gram elimination alone yields R/A gains of 1.02/1.03\(\times\) on RTX 5070Ti and 1.05/1.07\(\times\) on Ascend 910C. Adding \sysname{} with Gram elimination fixed yields 1.05/1.07\(\times\) and 1.07/1.09\(\times\), respectively. These paired controls establish a producer benefit beyond the companion; standalone measurements from other runs are not multiplied to infer a conditional gain. Component rows with $\dagger$ retain their measured-version qualification.

\paragraph{Component allocation effects.}
\begin{figure}[htbp]
\centering
\includegraphics[width=\linewidth]{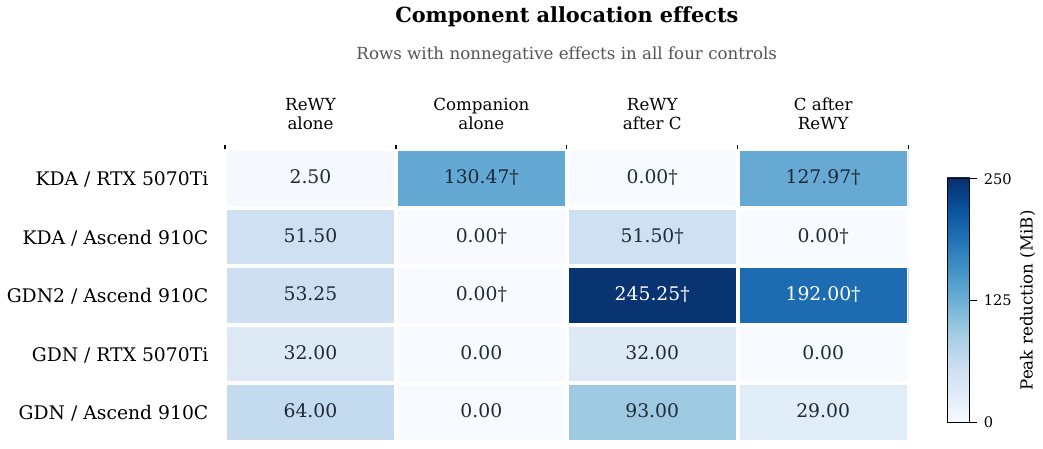}
\Description{A heat map shows component memory reductions in MiB for five operator-device pairs whose four recorded effects are nonnegative. Columns distinguish each component alone and each component after its companion. Positive values indicate savings; daggers mark historical precision limitations.}
\caption{Component changes in incremental peak allocation at \(B=1,T=8192,H=16,d_k=d_v=128\). Only operator/device rows with nonnegative effects in all four controls are displayed. C is matrix intra for KDA/GDN2 and Gram elimination for GDN. R/A values agree at the displayed precision. Dagger-marked cells are historical GPU and Ascend 910C matrix-consumer measurements. Table~\ref{tab:component_space} gives the schedules and exact values.}
\label{fig:component_memory}
\end{figure}

Figure~\ref{fig:component_memory} separates measured allocation changes. GDN/NPU's residual-only control saves 64\,MiB, with another 29\,MiB saved by adding Gram elimination. The allocation effect depends on the overlap of shared sensitivity and downstream-gradient lifetimes. Tables~\ref{tab:space} and~\ref{tab:component_space} give the complete peaks and component effects; a component ablation is not an additional production version.

\subsection{Numerical validation}
Validation for the original evaluation compares outputs, final states, and every input gradient. The separate 950 coverage is specified in Section~\ref{sec:ascend950}. An independent CPU FP64 recurrence supplies references through \(T=2048\); the \(T=8192\) workloads compare complete gradients with the frozen baseline. Relative \(L_2\) thresholds are 0.8\% for ordinary outputs/gradients and 2\% for gate, beta, and erase/write gradients. Near-zero references use a maximum absolute threshold of \(10^{-6}\), and all values must be finite.

The evaluated implementations pass the main workloads and the tested extended interfaces. The KDA fallback has a bitwise-determinism limitation described in Appendix~\ref{app:kda_determinism_limit}. Numerical passes do not establish the excluded paths' compliance with the precision contract. These operator checks are separate from the model-level diagnostics in Section~\ref{sec:e2e}. The suite includes tail chunks, unequal widths, nonzero state boundaries, zero/tiny beta, strong decay, and independent GDN2 gates. Graph replay uses intervening inputs to check outputs and temporary-buffer reuse. Appendix~\ref{app:evidence} specifies input distributions and additional BF16, FP16, and layout coverage.

\FloatBarrier
\subsection{Ascend 950DT: public autograd}
\label{sec:ascend950}
We evaluate the same four operators on one Ascend 950DT with 96\,GiB HBM and 248\,KiB usable UB. The software stack is CANN 9.2.0, PyTorch 2.10.0, torch\_npu 2.10.0.post7, and Triton-Ascend 3.6.0. The six shapes in Eq.~\eqref{main:shapes}, BF16 inputs, $C=64$, and $d_k=d_v=128$ are unchanged. Existing FP32 working matrices and accumulations are preserved. KDA and GDN2 use FP32 matrix intra with uniform $32\times64$ working tiles, distinct from the chunk and head dimensions.

\paragraph{Portability and the reference.}
The frozen FLA revision is unchanged, with uniform runtime adaptations required by this device: original GDN/GDN2 WY feature tiles are capped at 64 to fit UB, and three original KDA/GDN2 WY kernels use one program per task. The latter change removes repeat-dependent gradient errors observed with their persistent task loops. DeltaNet's residual producer requires the same scheduling change. Arithmetic-AST checks confirm that all four replacements preserve the original arithmetic after removing only the outer task loop; independent CPU equations additionally check the three original WY replacements. The precise compiler-level cause has not been established. All relevant comparisons use this corrected reference without workload-specific dispatch.

\paragraph{Protocol and acceptance.}
We time synchronized eager execution through public autograd, including allocations and gradient conversions. B excludes forward execution; FB includes both passes. These ratios are separate from low-level graph measurements and do not measure a complete training step. Thirty alternating ABBA/BAAB rounds yield 60 observations per implementation after at least six warm-up rounds and two seconds of warm-up. Ratios use per-shape medians and their equal-weight geometric mean across six shapes. KDA and GDN2 expose both policies. The public APIs of GDN and DeltaNet support R only.

The combined implementation passes 12 complete checks of forward and backward against a CPU FP64 recurrence at $T=64$. Each standalone branch passes eight corresponding checks. Every phase also passes ten actual-FLA-input matrix tests, including wide gates and batch two. Across all six shapes, 240 full-gradient/output/state checks pass, including 120 exact same-input repeat comparisons of original and optimized branches. Each timing case rechecks every gradient and optimization dispatch. Thresholds remain 0.8\% for ordinary quantities and 2\% for gate/beta gradients, with the same near-zero rule. The 950 suite does not establish FP16, variable-length, graph-replay, or model-training convergence coverage.

\begin{table}[htbp]
\footnotesize
\centering
\caption{Ascend 950DT combined paths: public-interface six-shape speedups. KDA/GDN2 combine \sysname{} and FP32 matrix intra; GDN uses residual/Gram elimination, and DeltaNet uses the unified chunk adjoint.}
\label{tab:950combined}
\begin{tabular}{llrr}
\toprule
Operator & Policy & B & FB\\
\midrule
KDA & R & 1.2177 & 1.1057\\
KDA & A & 1.3031 & 1.1171\\
GDN & R & 1.2192 & 1.1098\\
GDN2 & R & 2.4301 & 1.6432\\
GDN2 & A & 2.8945 & 1.7260\\
DeltaNet & R & 1.0878 & 1.0693\\
\bottomrule
\end{tabular}
\end{table}

\begin{table}[htbp]
\footnotesize
\centering
\caption{Independent 950 controls, each paired with its own same-policy original reference. These are standalone speedups, not conditional increments.}
\label{tab:950independent}
\begin{tabular}{lllrr}
\toprule
Component & Operator & Policy & B & FB\\
\midrule
\sysname{} & KDA & R & 1.0501 & 1.0158\\
\sysname{} & KDA & A & 1.0663 & 1.0190\\
\sysname{} & GDN2 & R & 1.1631 & 1.0849\\
\sysname{} & GDN2 & A & 1.1874 & 1.0915\\
FP32 matrix intra & KDA & R & 1.1417 & 1.0652\\
FP32 matrix intra & KDA & A & 1.1836 & 1.0724\\
FP32 matrix intra & GDN2 & R & 1.8385 & 1.4385\\
FP32 matrix intra & GDN2 & A & 2.0396 & 1.4820\\
\bottomrule
\end{tabular}
\end{table}

GDN2 has the largest combined gain, while DeltaNet remains below 1.1$\times$ and KDA's standalone residual gain is modest. The independent FP32 matrix controls demonstrate a substantial contribution for GDN2 under the stated storage contract. Combined, residual-only, and matrix-only ratios are directly measured against their respective references; we do not divide them to infer conditional gains. No new saving capability or 950 memory result is claimed. These operator results do not establish an end-to-end training benefit.
\FloatBarrier

\subsection{H100 and H200 backward evaluation}
\label{sec:h100}
We evaluate all four operators on NVIDIA H100 80GB HBM3 and NVIDIA H200 (both SM90), using PyTorch 2.12.1+cu126 and Triton 3.7.1. We use the six shapes in Eq.~\eqref{main:shapes}, $C=64$, and $d_k=d_v=128$, with the same frozen FLA reference and matched R/A policies. CUDA Graph device events yield 60 observations per implementation and shape. Table~\ref{tab:h100_operator} reports geometric means of six ratios of median complete-backward times. This boundary includes required recomputation and excludes the public wrapper's final gradient-type conversion.

\begin{table}[htbp]
\footnotesize\centering
\caption{Hopper complete low-level backward speedups ($\times$). Each entry is a six-shape geometric mean under matched recomputation (R) or full saving (A).}
\label{tab:h100_operator}
\setlength{\tabcolsep}{7pt}
\begin{tabular}{@{}lrrrr@{}}
\toprule
& \multicolumn{2}{c}{H100} & \multicolumn{2}{c}{H200}\\
\cmidrule(lr){2-3}\cmidrule(lr){4-5}
Operator & R & A & R & A\\
\midrule
DeltaNet & 1.2066 & 1.3046 & 1.1776 & 1.2697\\
GDN & 1.2056 & 1.2829 & 1.1819 & 1.2505\\
KDA & 1.3264 & 1.4213 & 1.3065 & 1.3964\\
GDN2 & 1.5680 & 1.7033 & 1.5408 & 1.6527\\
\bottomrule
\end{tabular}
\end{table}

KDA combines \sysname{} with FP32 endpoint-scaled matrix intra in a separate schedule. GDN2 uses an FP32 matrix-intra joint consumer and state tiles spanning the full key width and 16 value columns. GDN combines the unified adjoint, Gram elimination, and a liveness-aware joint consumer. DeltaNet uses the unified adjoint and state tiles of key width 64 and value width 16. All retain existing kernel arithmetic and FP32 precision.

All 48 operator/shape/policy combinations per device accelerate. Checks against CPU FP64, complete baseline gradients, and graph replay pass. Gains reflect the complete implementations; Hopper component ablations are not available. Software stacks differ across device groups, so differences in speedup cannot be attributed solely to hardware or scheduling. Section~\ref{sec:e2e} reports full-model backward; Appendix~\ref{app:h100_evidence} summarizes operator validation.

\subsection{Supplementary full-model backward}
\label{sec:e2e}
We integrate the optimized cores into existing FLA language models and measure the complete model backward pass. KDA uses FLA's \texttt{KDA\allowbreak For\allowbreak CausalLM}; GDN and DeltaNet use its native CausalLM classes with published structures~\cite{fla_models,gdn_config,delta_config}. Each model includes its original projections, convolutions, normalization, MLPs, language-model head, and loss. Only the core implementation changes within each matched comparison.

GDN2 uses the native \texttt{Gated\allowbreak Delta\allowbreak Net2} layer in FLA's common GDN language-model shell, retaining the official 1.3B model's key dimensions~\cite{fla_gdn2,gdn2_code}. It is a \emph{FLA layer port}, not the unchanged official full model. Appendix~\ref{app:e2e} specifies configurations and numerical limitations. These are not full Kimi Linear or Qwen3-Next deployments.

\paragraph{Backward-only boundary.}
All full-model tables report B, the complete \texttt{loss.\allowbreak backward()} interval. Forward/loss evaluation, gradient clearing, optimizer execution, compilation, initialization, and data loading are outside this measured interval. Backward includes every model layer and required recomputation, rather than only the substituted attention core. Both sides use native recomputation R, no layer checkpointing, BF16 autocast, and FP32 parameters, gradients, existing working matrices, and accumulations. Models are randomly initialized and inputs are resident synthetic tokens. The reported acceleration percentage is
\begin{equation}
 G_{\rm B}(\%)=100\left(
 \frac{\operatorname{median}(t_{\rm B,original})}
      {\operatorname{median}(t_{\rm B,optimized})}-1\right).
 \label{eq:e2e_backward}
\end{equation}
The tables show positive backward gains. This percentage measures increased backward execution rate, rather than percentage latency reduction or training throughput. For a group of configurations, we first take the geometric mean of all its original/optimized median-time ratios, then subtract one and multiply by 100; we do not average percentages or filter samples within a group. Original and final optimized full models are compared; component ablations remain at the operator level.

\paragraph{A800.}
CUDA events bracket backward inside a captured updating graph. Four ABBA blocks yield 60 observations per implementation; forward and optimizer work execute outside the event interval. Table~\ref{tab:e2e} reports the resulting backward times and gains.

\begin{table}[htbp]
\footnotesize
\centering\setlength{\tabcolsep}{4pt}
\caption{A800 full-model backward cases with positive gains, batch size 1. Times are CUDA-event B intervals. Numerical diagnostics are reported in Appendix~\ref{app:e2e_numerics}.}
\label{tab:e2e}
\begin{tabular}{@{}lrrrr@{}}
\toprule
Model & $T$ & \shortstack{Original B\\(ms)} & \shortstack{Optimized B\\(ms)} & \shortstack{B gain\\(\%)}\\
\midrule
KDA & 2048 & 112.547 & 103.068 & $+9.20$\\
KDA & 8192 & 358.533 & 325.021 & $+10.31$\\
GDN & 8192 & 334.440 & 333.498 & $+0.28$\\
DeltaNet & 2048 & 87.883 & 87.056 & $+0.95$\\
DeltaNet & 8192 & 277.062 & 274.461 & $+0.95$\\
GDN2 port & 2048 & 110.450 & 102.344 & $+7.92$\\
GDN2 port & 8192 & 371.798 & 342.511 & $+8.55$\\
\bottomrule
\end{tabular}
\end{table}

Across KDA's eight fixed-architecture input-size points, the aggregate B acceleration is 10.04\%. The displayed GDN case improves by 0.28\%, DeltaNet by about 0.95\%, and the GDN2 port by 7.92--8.55\%. Backward work outside the substituted core limits the full-model gain.

\paragraph{H100 and H200.}
Using the stacks in Section~\ref{sec:h100}, we measure backward event intervals in continuously updating graphs. Four fresh-model ABBA blocks yield 60 observations per implementation and configuration; forward and optimizer work execute outside the measured interval. Table~\ref{tab:e2e_h100} shows fixed-architecture groups with positive aggregate gains on both devices. Each group includes all its input configurations and uses a fixed backend. Each percentage transforms the geometric-mean ratio as in Eq.~\eqref{eq:e2e_backward}. Comparisons that change architecture or loss workload are excluded.

\begin{table}[htbp]
\footnotesize\centering\setlength{\tabcolsep}{3pt}
\caption{Hopper full-model pure-backward acceleration (\%): groups with positive aggregate gains on both devices. Native R, no layer checkpointing. GDN2 retains official dimensions in a FLA layer port.}
\label{tab:e2e_h100}
\begin{tabular}{@{}lrrr@{}}
\toprule
Model / scope / backend & Points & H100 & H200\\
\midrule
KDA / default & 8 & $+9.93$ & $+8.83$\\
GDN / published / wide64 & 2 & $+4.48$ & $+4.44$\\
DeltaNet / published & 2 & $+1.89$ & $+1.24$\\
GDN2 port / official dims. & 2 & $+6.51$ & $+5.34$\\
\bottomrule
\end{tabular}
\end{table}

Full-model backward includes projections, MLPs, normalization, and other work; its shapes and gate distributions also differ from the standalone sweep. These measurements cannot isolate the causes of the smaller model gains. The tested configurations pass execution and finite-value checks. KDA also has local and trajectory diagnostics. The other families are performance-only; some DeltaNet trajectory diagnostics exceed the acceptance thresholds. Operator passes do not establish full-model numerical acceptance or training convergence (Appendix~\ref{app:h100_models}).

\paragraph{Ascend 910C}
We use the frozen FLA Triton-Ascend reference with the same necessary platform adaptations on both sides. B uses synchronized eager wall-clock timing after forward has completed and synchronized; it excludes optimizer work. Four ABBA blocks yield 30 B samples per implementation. All models use $B=1$ and $T=2048$, with the architecture dimensions in Table~\ref{tab:model_provenance}. KDA and GDN2 combine \sysname{} with FP32 matrix intra; GDN and DeltaNet use their complete implementations. These measurements have a different timing method from A800, so we do not pool cross-device ratios or compare absolute latencies as hardware throughput.

\begin{table}[htbp]
\footnotesize
\centering\setlength{\tabcolsep}{7pt}
\caption{Ascend 910C full-model backward at $B=1,T=2048$, synchronized eager B. These are performance-only records; full-model gradient differences exceed some original thresholds. Each percentage compares runs on the same physical device.}
\label{tab:e2e_910}
\begin{tabular}{lrrr}
\toprule
Model & Original B (ms) & Optimized B (ms) & B gain (\%)\\
\midrule
KDA & 204.359 & 187.414 & $+9.04$\\
GDN & 175.469 & 162.777 & $+7.80$\\
DeltaNet & 169.257 & 139.724 & $+21.14$\\
GDN2 port & 226.075 & 158.905 & $+42.27$\\
\bottomrule
\end{tabular}
\end{table}

\paragraph{Ascend 950DT coverage.}
The Ascend 950DT experiments in Section~\ref{sec:ascend950} time complete operators, not a complete model. No full-model B measurement is available for that device; no full-model speedup is inferred from its operator ratios.

\paragraph{Numerical scope.}
KDA on A800 passes layer-VJP tests, graph and eager equivalence, and finite-update checks. Full-network gradient differences remain diagnostics. Other A800 and Ascend 910C models pass interface, finite-value, backend-activation, and execution checks. Appendix~\ref{app:e2e_numerics} reports gradient and parameter differences.
\FloatBarrier

\section{Discussion and Conclusion}
\label{main:conclusion}
\sysname{} shows that efficient delta-rule training benefits from simplifying the derivative graph before optimizing its kernels. A shared reverse sensitivity removes the inverse adjoint and reduces seven local products to three. Matrix intra and Gram elimination expose further reuse in vector- and scalar-gated interactions. These algebraic identities preserve the existing recurrence and apply to four operator parameterizations already supported by community implementations.

Complete-backward measurements and paired controls establish the operator-level benefit. Allocation results show how intermediate lifetimes affect the mapping of the reduced graph to hardware. Supplementary FLA integrations measure the smaller, workload-dependent gains in full-model backward. Exact arithmetic equivalence does not guarantee identical floating-point trajectories. We therefore distinguish operator validation, historical precision-contract violations, and model-level numerical diagnostics. The contribution reduces the work of an existing linear-attention primitive in a form that accelerator kernels can implement.
\label{main:end}

\phantomsection
\label{references:start}
\bibliographystyle{ACM-Reference-Format}
\bibliography{misc/references}

\label{references:end}
\clearpage
\appendix
\section{Derivation of the Chunk System}
\label{app:forward}
For the recurrence in Section~\ref{main:problem}, unrolling the state entering token \(i\)'s erase operation gives
\begin{equation*}
 S_i^-=\diag(\gamma_i)S+
 \sum_{j<i}\diag(\gamma_i\invHad\gamma_j)k_jr_j^\top.
\end{equation*}
Here \(\gamma_i=\exp(\sum_{t\le i}g_t)\); a write at \(j\) receives decays from \(j+1\) through \(i\). With the row-stacked factors
\begin{equation}
 \begin{aligned}
 E&=B_e\Had K\Had\gamma,&Y&=K\invHad\gamma,\\
 Z&=W_e\Had V,&Q_g&=Q\Had\gamma,
 \end{aligned}
 \label{eq:factors}
\end{equation}
substitution yields \(R_i=Z_i-E_iS-\sum_{j<i}(E_iY_j^\top)R_j\).
Stacking rows gives \(MR=Z-ES\), with \(M=I+\low(EY^\top)\).
This unit lower triangular system is nonsingular for all compatible dimensions: its strict lower part satisfies \(L^n=0\).
Forward substitution recovers each correction and therefore the original state sequence. The argument also applies to tail chunks.

Let \(\Lambda=\diag(\gamma_n)\), \(F=Y\Lambda\), and \(C_q=\causal(Q_gY^\top)\). Unrolling the read and final state gives
\begin{equation}
 O=Q_gS+C_qR,\qquad S^+=\Lambda S+F^\top R.
 \label{eq:outputs}
\end{equation}
The output includes the current write, so \(C_q\) has an inclusive mask; the erase system uses a strict mask. The existing forward organization \(W=AE\), \(U=AZ\), \(R=U-WS\), with \(A=M^{-1}\), is retained.

\section{Complete Gradients and Equivalence}
\label{sec:residual}
\subsection{Residual pullback}
\label{app:residual_proof}
\begin{proof}[Proof of Theorem~\ref{thm:residual}]
Differentiating \(MR=Z-ES\) gives
\begin{equation*}
 M\,\mathrm dR=\mathrm dZ-(\mathrm dE)S-E\,\mathrm dS-(\mathrm dL)R.
\end{equation*}
Since \(M^\top P=\bar R\), the Frobenius pairing yields
\begin{equation*}
 \begin{aligned}
 \mathrm d\ell&=\ip{P}{\mathrm dZ}-\ip{PS^\top}{\mathrm dE}\\
 &\quad-\ip{E^\top P}{\mathrm dS}-\ip{PR^\top}{\mathrm dL}.
 \end{aligned}
\end{equation*}
Allowed variations of \(L\) are strictly lower triangular, giving \(D=-\low(PR^\top)\).
The mask is self-adjoint. Substituting
\(\mathrm dL=\low((\mathrm dE)Y^\top+E(\mathrm dY)^\top)\)
gives the remaining contributions \(DY\) and \(D^\top E\).
\end{proof}

\paragraph{Cancellation in the expanded graph.}
\label{app:expanded_wy}
The conventional graph has \(\bar W=-\bar RS^\top\) and
\(\bar A=\bar RZ^\top+\bar WE^\top=\bar R(Z-ES)^\top\).
Thus its inverse pullback simplifies to
\begin{equation*}
 A^\top\bar A A^\top
 =(A^\top\bar R)[A(Z-ES)]^\top=PR^\top,
\end{equation*}
and \(A^\top\bar W=-PS^\top\). This removes the two auxiliary adjoints without changing any required gradient.

\subsection{Output, state, and effective-input gradients}
\label{sec:complete}
The incoming \(\bar S^+\) includes later chunks and any terminal-state adjoint. Differentiating Eq.~\eqref{eq:outputs} gives the complete map
\begin{align*}
 \bar R&=C_q^\top\bar O+F\bar S^+,&H_q&=\causal(\bar OR^\top),\\
 \bar F&=R(\bar S^+)^\top,&\bar Q_g&=\bar OS^\top+H_qY,\\
 \bar E&=-PS^\top+DY,\\
 \bar Y&=H_q^\top Q_g+D^\top E+\bar F\Lambda,\\
 \bar S&=Q_g^\top\bar O+\Lambda\bar S^+-E^\top P.
\end{align*}
The scan already supplies \(-E^\top P=-W^\top\bar R\); it must be accumulated once, not added again by the local consumer. The chain rule for Eq.~\eqref{eq:factors} gives
\begin{align*}
 \bar Q&=\gamma\Had\bar Q_g,\\
 \bar K&=\gamma\Had B_e\Had\bar E+\gamma^{-1}\Had\bar Y,\\
 \bar B_e&=\gamma\Had K\Had\bar E,\\
 \bar V&=W_e\Had P,&\bar W_e&=V\Had P.
\end{align*}
GDN2 returns independent erase/write gradients. With \(\lambda=\gamma_n\), the endpoint contribution is
\begin{equation*}
 h_k=\lambda_k\left[
 \sum_v S_{kv}\bar S^+_{kv}+\sum_iY_{ik}\bar F_{ik}\right].
\end{equation*}
The cumulative and token-local decay derivatives are
\begin{equation*}
 \begin{aligned}
 \bar G_i&=(Q_g)_i\Had(\bar Q_g)_i+E_i\Had\bar E_i\\
 &\quad-Y_i\Had\bar Y_i+\mathbf1_{\{i=n\}}h^\top,\\
 \bar g_t&=\sum_{i=t}^{n}\bar G_i.
 \end{aligned}
\end{equation*}
Scalar decay sums over features. The endpoint differentiates \(\lambda\); the term in \(\bar Y\) differentiates the token-local reciprocal factor.

\subsection{Shared beta and scalar decay}
For shared erase/write beta, set \(X=\gamma\Had K\).
The complete, undivided derivative is
\begin{equation}
 \bar\beta_i=\langle X_i,\bar E_i\rangle+\langle V_i,P_i\rangle.
 \label{eq:betadirect}
\end{equation}
\begin{corollary}[Shared-beta contraction]
\label{cor:beta}
\(\beta_i\bar\beta_i=\langle P_i,R_i\rangle\).
\end{corollary}
\begin{proof}
Multiplying Eq.~\eqref{eq:betadirect} by \(\beta_i\) gives
\begin{align*}
 \beta_i\bar\beta_i
 &=\langle P_i,Z_i-E_iS\rangle+
   \sum_{j<i}D_{ij}L_{ij}\\
 &=\langle P_i,(MR)_i\rangle-
   \sum_{j<i}\langle P_i,R_j\rangle L_{ij}\\
 &=\langle P_i,R_i\rangle.
\end{align*}
\end{proof}
The quotient is used only for sufficiently nonzero beta. At zero or tiny beta the undivided expression remains defined.

For scalar GDN, the erase-only contribution is
\(b_i^{\rm erase}=\langle\gamma_iK_i,\bar E_i\rangle\).
Its token-local log-gate derivative can be computed from completed FP32 accumulators:
\begin{equation*}
 c_i=\langle\bar Q_i,Q_i\rangle-\langle\bar K_i,K_i\rangle
       +2\beta_i b_i^{\rm erase}.
\end{equation*}
This equals the feature sum of the first three terms in \(\bar G_i\), including at zero beta.
The state endpoint is \(\lambda\ip{S}{\bar S^+}\);
the key endpoint is \(\lambda\sum_i\langle Y_i,\bar F_i\rangle\).
Both are added at the last valid token before the reverse sum.
The write-beta path is independent of the log gate. The forward base-2 conversion retains the natural-log chain rule; no additional \(\ln2\) factor is introduced.

\subsection{Arithmetic accounting}
\label{app:residual_cost}
Counting a multiply and add as two operations, the seven expanded products and three reduced products cost
\begin{align*}
 \mathcal F_{\rm expanded}
 &=2nd_kd_v+4n^2d_v+4n^2d_k+4n^3,\\
 \mathcal F_{\rm residual}&=2nd_kd_v+4n^2d_v.
\end{align*}
The difference is \(4n^2d_k+4n^3\), excluding common interactions, the state scan, and recomputation. Triangular-block skipping or repeated production of \(P\) changes the executed count, not the identity.

\section{Matrix-Intra Equivalence and Centering}
\label{sec:intra}
For fixed coefficient \(C\) and cumulative gates \(G\), define
\begin{equation*}
 [\mathcal T_{C,G}(X)]_{ik}
 =\sum_j C_{ij}X_{jk}\exp(G_{ik}-G_{jk}).
\end{equation*}
With \(\mathcal D_G(X)=\exp(G)\Had X\) and \(\mathcal M_C(X)=CX\),
\begin{equation*}
 \mathcal T_{C,G}=\mathcal D_G\circ\mathcal M_C\circ\mathcal D_{-G}.
\end{equation*}
Expanding the finite sum proves the factorization. The scaling maps are self-adjoint and \(\mathcal M_C^*=\mathcal M_{C^\top}\), hence
\(\mathcal T_{C,G}^*=\mathcal T_{C^\top,-G}\).
These identities yield the contractions in Eq.~\eqref{main:intra} and gradients in Eq.~\eqref{main:intraoutputs}, preserving both causal masks.

For Eq.~\eqref{main:diagonalaction}, diagonal factors cancel in
\(\widetilde E\widetilde Y^\top=EY^\top\) and
\(\widetilde Q_g\widetilde Y^\top=Q_gY^\top\).
The contractions transform as
\begin{equation*}
 \widetilde U_q=U_qC_c,\quad
 \widetilde U_e=U_eC_c,\quad
 \widetilde U_y=U_yC_c^{-1}.
\end{equation*}
Replacing \(\gamma,\gamma^{-1}\) by \(\exp(G-c),\exp(c-G)\) cancels these factors in every physical gradient.
This holds for any center, including a data-dependent center: it is an invariant execution representation, not a new differentiated operator.
Direct-state products retain their original coordinates.

For any feature center \(c_k\), the triangle inequality gives
\(\max(|M_k-c_k|,|m_k-c_k|)\ge(M_k-m_k)/2\).
The midpoint attains equality, proving the bound in Section~\ref{main:matrixintra}. A tile center uses the extrema over the tile and gives the same conservative bound.
Range/layout checks select a validated pairwise or subchunk fallback; the recorded GDN2 threshold is 80 in log-base-2 units.
Feature tiling selects columns and does not alter any token contraction.

\section{Gram-Free Reduction Proof}
\label{sec:gram}
For \(X\in\RR^{m\times d}\), \(Y\in\RR^{n\times d}\), and
\(C\in\RR^{m\times n}\), let \(J=C\Had(XY^\top)\).
Finite-sum interchange gives
\begin{align*}
 \sum_jJ_{ij}&=\sum_k X_{ik}(CY)_{ik},\\
 \sum_iJ_{ij}&=\sum_k Y_{jk}(C^\top X)_{jk}.
\end{align*}
No symmetry, sign, or mask assumption on \(C\) is needed.
Taking \(X=Y=K\) and \(C=\diag(\beta)\Gamma\) proves the scalar-gate identity in Section~\ref{main:gram}:
\begin{equation*}
 \begin{aligned}
 \sum_jJ_{ij}&=\beta_i\langle K_i,(A_K)_i\rangle,\\
 \sum_jJ_{ji}&=\langle K_i,(B_K)_i\rangle.
 \end{aligned}
\end{equation*}
Subtracting the two reductions gives \(\bar G_i^{(L)}\).
The erase-beta contribution is \(\langle K_i,(A_K)_i\rangle\), valid at zero beta.
Direct contributions are added before shared-beta accumulation and the reverse gate sum.

Elimination requires that consumers need only these reductions and that the feature contractions are already computed. A consumer of the full pairwise matrix prevents it.
Vector-gated KDA/GDN2 generally have no single scalar \(\Gamma_{ij}\) and instead use matrix intra. Reordered floating-point reductions are subject to the same complete-gradient checks, not an exact floating-point equivalence claim.

\section{Complete Reverse Execution and Boundaries}
\label{app:algorithm}
The complete reverse algorithm traverses chunks in reverse order:
\begin{enumerate}
\item Accumulate \(\bar R\) from both output and outgoing-state paths; form \(H_q\) and \(\bar F\).
\item Apply Algorithm~\ref{alg:residual}, then add the output and final-state contributions to \(\bar Q_g,\bar Y,\bar S\).
\item Apply the factor chain rules in Appendix~\ref{sec:complete}; sum erase/write contributions for shared beta.
\item Add both decay endpoints at the last valid token and reverse-sum \(\bar G\). Restore prescribed query scaling, normalization, and outer gate derivatives.
\end{enumerate}
The scan supplies the incoming-state term once.
Causal masking prevents arbitrary reassociation. In general,
\begin{equation*}
 \low(PR^\top)Y\ne P(R^\top Y).
\end{equation*}
Reassociation is valid within complete blocks below the diagonal; diagonal blocks retain the strict mask.

\subsection{Partitions and shared parameters}
\label{app:generality}
Each valid chunk length has its own unit triangular system. Reverse composition therefore handles nonuniform partitions and tail chunks without a new identity.
Feature tiles preserve finite contractions.
For a broadcast gate \(B_e=\mathcal B\theta\), the adjoint is
\(\bar\theta=\mathcal B^*\bar B_e\):
scalar gates sum features, shared beta sums erase/write paths, and grouped heads sum shared-input contributions.

If only fraction \(f\) of baseline time changes with local speedup \(s\), the ideal complete-operator ratio is \(1/(1-f+f/s)\). The state scan, recomputation, transfers, and launches limit the measured gain; local product counts alone do not predict it.

\section{Schedules and Numerical Contract}
\label{app:execution}
Joint consumers reduce transfers but extend live ranges; separate consumers shorten them and add scratch transfers.
RTX 5070Ti KDA streams value tiles of width 64; its state scan maintains value groups of width 16 in FP32.
A800 DeltaNet uses the original K64 state arithmetic with three stages, while GDN retains the original state autotuner. Their scalar consumers finish FP32 value/weight uses of \(P\) before dependent temporaries, avoiding a global \(P\) buffer.
The references retain native autotuning and candidates keep fixed settings across shapes.

On Ascend 910C, GDN stores weighted residual and output coefficients in FP32. The matrix consumer computes query and key gradients, the erase-beta reduction, and the key endpoint. A separate Vector kernel adds the state endpoint and reverse-sums the gate gradient.
This adds one FP32 scalar per token/head and one kernel boundary; complete timing includes both stages and transfers.
Hopper and Ascend 950DT schedules are specified with their results in Sections~\ref{sec:h100} and~\ref{sec:ascend950}.

\begin{table}[t]
\centering\small
\caption{Retained storage and accumulation contracts. Measured-version exclusions are in Appendix~\ref{app:precision_status}.}
\label{tab:gpu_precision}
\label{tab:gdn_precision}
\begin{tabularx}{\linewidth}{>{\raggedright\arraybackslash}p{.29\linewidth}>{\raggedright\arraybackslash}X}
\toprule
Quantity & Contract\\
\midrule
State scan & FP32 accumulators and initial-state gradient; stored chunk states follow the reference format.\\
Shared \(P\) & RTX scalar/KDA paths store the input format after required FP32 uses; GDN/Ascend 910C retains FP32 for value/beta consumers.\\
Existing matrices & Eligible paths retain FP32 \(L/D\) arrays and prescribed GEMM conversions. Output casts cannot repair a lower internal operand precision.\\
GDN/Ascend 910C & Weighted coefficients and key contractions remain FP32. Gate scalars and both endpoints are reduced before output rounding.\\
\bottomrule
\end{tabularx}
\end{table}

Reordered products change floating-point rounding. Centering and range checks constrain exponential magnitudes but do not bound the triangular system's condition number.
No general floating-point stability theorem is claimed. Zero/tiny beta uses the undivided derivative; GDN2 retains independent erase/write gradients.

\section{Protocol, Precision, and Validation}
\label{app:evidence}
\subsection{Frozen controls and measurement scope}
The reference is FLA revision \nolinkurl{a59bc5df8c5841dce2169bb00e4027287197ba77}.
GPU controls are original. Ascend 910C adaptations retain the equations: an ungated DeltaNet schedule, stable independent key/transpose loads for GDN, and original GDN2 intra equations.
The fragmented ungated DeltaNet reference duplicates work under a transpose-transfer compiler limitation and contributes to its larger gain.
KDA uses its existing saving switch; other matched wrappers cache identical intermediates.
One implementation is fixed per device/operator across six shapes and matched R/A policies.

Original GPU graph runs use six extra warm-up seconds and 30 balanced rounds (60 observations); Ascend 910C uses at least two seconds and ten rounds (20 observations).
Each graph contains two calls and times are normalized per call.
Per-shape medians are reduced by an equal-weight geometric mean of reference/optimized ratios.
Gradients are checked before and after timing and on replay with intervening inputs.
The Ascend 950DT public eager boundary, adaptations, and acceptance are separately specified in Section~\ref{sec:ascend950}; the Hopper protocol is in Section~\ref{sec:h100}.
These cohorts are not pooled.

\subsection{Precision eligibility}
\label{app:precision_status}
The dagger-marked complete entries are A800 KDA/GDN2, RTX 5070Ti GDN2, and Ascend 910C KDA/GDN2.
Their measured matrix paths cast existing FP32 \(L/D\) operands to BF16.
The corresponding matrix-only and conditional controls, including the separate RTX KDA controls, have the same limitation.
They remain historical timing/allocation records, not evidence for current precision-qualified claims.
Residual-only controls and the RTX KDA joint measurement have separate eligibility.
Later numerical passes or public-output casts do not restore internal precision.

At \(B=1,T=2048,H=4\), historical conversion-inclusive KDA/GDN Ascend 910C controls give 1.27/1.34 and 1.41/1.61 R/A ratios, respectively. The KDA precision exclusion still applies.

\subsection{Numerical coverage}
The main harness uses feature-normalized Gaussian BF16 queries/keys, Gaussian values/output adjoints, query scale \(d_k^{-1/2}\), and FP32 Gaussian initial/terminal states scaled by 0.1.
Log gates are \(\log\sigma(x)/16\); erase/write weights lie in \([0.1,0.9)\).
CPU FP64 and complex-step checks cover multiple chunks, unequal widths, nonuniform partitions, nonzero state boundaries, shared gates, and both scalar endpoints.
Accelerator extensions include tails, zero/tiny beta, strong decay, and applicable packed/grouped/layout interfaces.
RTX KDA/GDN2 and A800 DeltaNet/GDN have recorded BF16 tests at widths 80/112 and 192/160 and FP16 tests at 96/80.
Coverage is limited to the tested interfaces; it does not remove precision exclusions or establish convergence.

\paragraph{KDA determinism limitation.}
\label{app:kda_determinism_limit}
An A800 test at \(T=65,d_k=d_v=64\), with every third beta zero, found beta-only repeat differences in 20 of 21 comparisons (maximum \(1.49\times10^{-8}\)).
The CPU tolerance check passed; the differences arise from unordered FP32 atomics in the fallback.
Thus tolerance-based correctness does not imply general bitwise determinism.

\subsection{Hopper validation}
\label{app:h100_evidence}
On each of H100 and H200, validation covers all four operators, six shapes, and both saving policies.
Checks against CPU FP64 references and complete baseline gradients pass, as do graph-replay checks with intervening inputs.
The implementations retain existing FP32 working matrices and accumulations.
These operator checks do not establish full-model numerical acceptance or training convergence.

\section{Allocation Definition and Controls}
\label{app:space}
Incremental peaks include warmed backward tensors and returned gradients.
Forward auxiliaries exclude inputs and forward outputs and are counted once per underlying storage. Saved bytes match within each policy.
Resident workspace is separate: the historical KDA/Ascend 910C matrix path uses 16\,MiB per stream.
Tables~\ref{tab:space} and~\ref{tab:component_space} retain the complete and component allocation values.

The two audited shapes are \((1,2048,4)\) and \((1,8192,16)\) in \((B,T,H)\) notation. Each uses both policies and three repetitions.
Repeated peaks agree exactly in the recorded GPU controls, with complete-gradient checks.
Allocation peaks depend on the tested schedule. Separate component controls cannot be substituted for the complete implementation's allocation.

\section{Numerical Values for the Result Figures}
\label{app:figure_values}

The tables report measured baselines and component ablations. Appendix~\ref{app:precision_status} specifies precision exclusions. Each $\dagger$ marks the same qualification as in the corresponding figure.

Tables~\ref{tab:main}--\ref{tab:component_space} list the original-cohort values displayed in Figures~\ref{fig:backward_speedups}--\ref{fig:component_memory}. Supplementary operator values are in Tables~\ref{tab:950combined} and~\ref{tab:h100_operator}.

\begin{table}[!htbp]
\centering
\caption{Complete-backward speedups: six-shape geometric means, R/A. DeltaNet uses \sysname{}; GDN adds Gram elimination; KDA/GDN2 add matrix intra. A800 KDA/GDN2 and NPU GDN2 use separate consumers; their RTX 5070Ti implementations use joint consumers. GPU GDN2, A800 KDA, and Ascend 910C KDA/GDN2 combined entries are historical, excluded from current claims (Section~\ref{app:precision_status}).}
\label{tab:main}
\footnotesize
\begin{tabular}{lccc}
\toprule
Operator & RTX 5070Ti & A800 & Ascend 910C\\
\midrule
DeltaNet &1.1469/1.2152&1.1474/1.2237&1.5806/1.7528\\
GDN &1.1614/1.2264&1.1732/1.2471&1.3739/1.5142\\
KDA &1.5560/1.8066&1.7943/2.0832&1.5464/1.6291\\
GDN2 &1.7910/2.1458&1.9577/2.2792&2.5518/2.9959\\
\bottomrule
\end{tabular}
\end{table}

\begin{table}[!htbp]
\centering
\caption{Standalone component B speedups, R/A, from separate runs against frozen references. The companion is matrix intra for KDA/GDN2 and Gram elimination for GDN. GPU matrix-companion entries are historical.}
\label{tab:ablation}
\footnotesize
\begin{tabular}{llcc}
\toprule
Operator & Device & \sysname{} only & Companion only\\
\midrule
KDA & RTX 5070Ti &1.0845/1.1160&1.1996/1.2600\\
KDA & NPU &1.1402/1.1660&1.3675/1.4179\\
GDN & RTX 5070Ti &1.0490/1.0852&1.0214/1.0323\\
GDN & NPU &1.0733/1.0785&1.0538/1.0658\\
GDN2 & RTX 5070Ti &1.1732/1.2026&1.2851/1.3407\\
GDN2 & NPU &1.1103/1.1217&2.0834/2.3268\\
\bottomrule
\end{tabular}
\end{table}

\begin{table}[!htbp]
\footnotesize
\centering
\caption{Paired gains from adding \sysname{} to a fixed companion. KDA/GDN2 RTX 5070Ti use separate residual/matrix consumers; KDA/NPU omits global direct-erase storage; GDN uses staged residual/Gram controls. GPU KDA/GDN2 entries are historical.}
\label{tab:conditional}
\begin{tabular}{llcc}
\toprule
Operator & Fixed companion & RTX 5070Ti R/A & NPU R/A\\
\midrule
KDA & Matrix intra &1.1070/1.1366&1.1308/1.1490\\
GDN & Gram elimination &1.0485/1.0663&1.0708/1.0889\\
GDN2 & Matrix intra &1.2364/1.2911&1.2248/1.2876\\
\bottomrule
\end{tabular}
\end{table}

\begin{table}[!htbp]
\centering
\caption{Incremental backward peak allocation, MiB, at \(B=1,T=8192,H=16,d_k=d_v=128\), R/A. Implementations match Table~\ref{tab:main}. KDA/NPU has an additional 16\,MiB resident workspace per stream. GPU GDN2 and A800 KDA entries are historical.}
\label{tab:space}
\footnotesize
\begin{tabular}{llrr}
\toprule
Operator & Device & Original & Optimized\\
\midrule
DeltaNet & RTX 5070Ti &417.250/257.250&417.250/257.250\\
DeltaNet & A800 &417.250/257.250&385.250/225.250\\
DeltaNet & NPU &610.508/450.505&518.383/358.380\\
GDN & RTX 5070Ti &418.250/258.250&417.750/257.750\\
GDN & A800 &418.250/258.250&385.750/225.750\\
GDN & NPU &612.510/452.508&520.636/360.634\\
GDN2 & RTX 5070Ti &961.000/737.000&833.000/609.000\\
GDN2 & A800 &961.000/737.000&801.000/577.000\\
GDN2 & NPU &1015.262/791.259&770.007/546.004\\
KDA & RTX 5070Ti &772.000/548.000&641.500/417.500\\
KDA & A800 &772.000/548.000&641.531/417.531\\
KDA & NPU &824.511/600.509&773.009/549.006\\
\bottomrule
\end{tabular}
\end{table}

\begin{table}[!htbp]
\centering
\caption{Component reductions in incremental peak allocation, MiB, at the same shape, R/A. Operator/device rows are shown when all four effects are nonnegative. O denotes \sysname{}; C is matrix intra for KDA/GDN2 and Gram elimination for GDN. KDA RTX 5070Ti uses separate consumers; GDN uses staged controls. GPU KDA companion and conditional entries are historical.}
\label{tab:component_space}
\footnotesize
\setlength{\tabcolsep}{0.6pt}
\begin{tabular}{@{}llrrrr@{}}
\toprule
Operator & Device & \shortstack{O\\alone} & \shortstack{C\\alone} & \shortstack{O after\\C} & \shortstack{C after\\O}\\
\midrule
KDA & RTX 5070Ti &2.50/2.50&130.47/130.47&0.00/0.00&127.97/127.97\\
KDA & NPU &51.50/51.50&0.00/0.00&51.50/51.50&0.00/0.00\\
GDN2 & NPU &53.25/53.25&0.00/0.00&245.25/245.25&192.00/192.00\\
GDN & RTX 5070Ti &32.00/32.00&0.00/0.00&32.00/32.00&0.00/0.00\\
GDN & NPU &64.00/64.00&0.00/0.00&93.00/93.00&29.00/29.00\\
\bottomrule
\end{tabular}
\end{table}

\section{Related Work}
\label{main:related}
\paragraph{Delta-rule computation.}
The WY representation originated as a compact representation of Householder products~\cite{wy}. DeltaNet uses related structure for parallel sequence computation. GDN adds adaptive forgetting, KDA channel-wise decay, and GDN2 independent erase and write vectors~\cite{deltanet,gdn,kda,gdn2}. Our transformations preserve these operators and their forward computations while simplifying the backward graph.

\paragraph{Matrix derivatives and kernel implementations.}
The adjoint of a linear solve is classical~\cite{giles}. \sysname{} applies it to the composite map \(MR=Z-ES\) and eliminates the inverse-adjoint chain while retaining all gate and state derivatives. FLA and its Ascend adaptation provide the chunk operators used as frozen references~\cite{fla,flanpu}. The audited FlashQLA revision includes a Hopper path that combines a residual right-hand side before forming the inverse adjoint, as well as scalar-key symmetrization~\cite{flashqla}. \sysname{} further removes the two-sided inverse pullback. cuLA provides hardware-specialized linear-attention kernels~\cite{cula}; a recent Blackwell study examines shared gated-recurrence backward computation with independent correctness verification~\cite{blackwell}.

\paragraph{Separable contractions.}
GLA derives scaled query and key factors and secondary chunking to balance matrix execution with numerical stability~\cite{gla}. KDA further exploits tied key factors to simplify its chunkwise computation~\cite{kda}. Our matrix-intra consumer uses these established structures in a whole-chunk backward organization with centering and explicit gate-gradient production; the separable exponential identity itself is not a novelty claim.

\paragraph{Intermediate elimination.}
FlashAttention reduces intermediate materialization and expresses a backward pairwise reduction as a feature-space inner product~\cite[Appendix~B.2]{flashattention}. \sysname{} eliminates intermediate adjoints in a triangular system. Matrix intra and Gram elimination reorganize its remaining gated contractions and reductions.

\section{Model Provenance and Numerical Scope}
\label{app:e2e}
The full models use the frozen FLA revision identified in Appendix~\ref{app:evidence}.
All have vocabulary 32,000; parameter counts include the complete instantiated language model.

\begin{table}[t]
\centering\footnotesize
\caption{Full-model dimensions. KDA uses FLA defaults; GDN/DeltaNet use published configurations. GDN2 is a FLA layer port, not the official full model.}
\label{tab:model_provenance}
\setlength{\tabcolsep}{3pt}
\begin{tabular}{lrrrrr}
\toprule
Model & Layers & Hidden & MLP & \(H/K/V\) & Params (B)\\
\midrule
KDA &24&2048&5632&16/128/128&1.391\\
GDN &24&2048&5632&8/256/256&1.466\\
DeltaNet &24&2048&5632&16/128/128&1.366\\
GDN2 port &18&2304&6208&16/128/128&1.450\\
\bottomrule
\end{tabular}
\end{table}

KDA's structure comes from FLA defaults~\cite{fla_models}; GDN and DeltaNet use their published dimensions~\cite{gdn_config,delta_config}.
The GDN2 adapter uses the native FLA layer~\cite{fla_gdn2} in a common language-model shell, with dimensions from the official model~\cite{gdn2_code}. Its sigmoid output gate and shell initialization differ from the NVlabs full model.
Matching dimensions is not architecture or initialization equivalence. Both compared cores use the same instantiated model; no row reproduces industrial pretrained weights.

\subsection{Execution and numerical diagnostics}
\label{app:e2e_numerics}
Runs use one device, native R, \(C=64\), and no gradient accumulation or layer checkpointing. Both sides use matched random initialization and inputs, BF16 autocast, and FP32 parameters, working matrices, and accumulations.
A800 and Hopper record 60 event intervals per implementation across four fresh-model ABBA blocks; Ascend 910C records 30 synchronized eager intervals.
Forward and optimizer work are excluded from B. Each ratio uses one physical device and a fixed backend; no case dispatcher selects the faster schedule.

KDA/A800 passes all 24 layer VJPs at the original thresholds: 0.8\% for ordinary gradients, 2\% for gates, and \(10^{-6}\) near zero.
With changing inputs, graph and eager executions agree elementwise within each algorithm.
Across eight fixed-architecture inputs, initial full-network gradient differences are 1.62--1.64\%; after 46 updates, parameter differences are 0.17--0.20\% and loss differences at most 0.22\%.
These diagnostics do not establish every parameter's accuracy or long-run convergence.

Ascend 910C maximum parameter-gradient differences are 1.94\% (KDA), 2.60\% (GDN), 2.25\% (DeltaNet), and 2.17\% (GDN2 port).
Some exceed the original thresholds; the model records are performance-only and remain qualified accordingly.
Operator correctness checks do not establish full-model numerical acceptance.

\subsection{Hopper configurations and numerical scope}
\label{app:h100_models}
KDA keeps its architecture fixed and uses eight \((B,T)\) pairs:
\begin{equation*}
 \begin{gathered}
 (1,2048),\quad (1,4096),\quad (2,2048),\quad (1,8192),\\
 (2,4096),\quad (4,2048),\quad (1,16384),\quad (4,4096).
 \end{gathered}
\end{equation*}

Hopper GDN2 uses the official-dimension port with \(B=1\) at \(T=2048\) and \(8192\).
Hopper GDN and DeltaNet use their published structures at the same two lengths; the separate GDN joint-consumer point uses $B=1,T=2048$. Model structure and backend remain fixed within each group.

All eight KDA input configurations have local-VJP, parameter, and loss diagnostics. GDN, GDN2, and DeltaNet measurements are performance-only: finite-value and execution checks do not establish full-model gradient acceptance. DeltaNet trajectory checks can exceed the thresholds.

Changing head counts, feature widths, or loss heads alters model capacity or workload. Such comparisons are excluded from the full-model tables, which keep architecture and loss fixed and report backward only. Operator ratios do not imply a model or complete-training gain.

\end{document}